\documentclass[conference]{IEEEtran}
\IEEEoverridecommandlockouts
\usepackage{cite}
\usepackage{amsmath,amssymb,amsfonts}
\usepackage{algorithmic}
\usepackage{graphicx}
\usepackage{textcomp}
\usepackage{xcolor}
\usepackage{color}
\usepackage{multirow}
\usepackage{pifont}
\def\BibTeX{{\rm B\kern-.05em{\sc i\kern-.025em b}\kern-.08em
    T\kern-.1667em\lower.7ex\hbox{E}\kern-.125emX}}
\begin{document}

\title{Understanding Depth Map Progressively: Adaptive Distance Interval Separation for Monocular 3d Object Detection}

\author{\IEEEauthorblockN{1\textsuperscript{st} Xianhui Cheng}
\IEEEauthorblockA{\textit{School of Computer Science} \\
\textit{Fudan University}\\
Shanghai, China \\
xianhuicheng20@fudan.edu.cn}

\and

\IEEEauthorblockN{2\textsuperscript{nd} Shoumeng Qiu}
\IEEEauthorblockA{\textit{School of Computer Science} \\
\textit{Fudan University}\\
Shanghai, China \\
smqiu21@m.fudan.edu.cn}

\and

\IEEEauthorblockN{3\textsuperscript{rd} Zhikang Zou}
\IEEEauthorblockA{\textit{Baidu Inc.} \\
Shenzhen, China \\
zouzhikang@baidu.com}

\and

\IEEEauthorblockN{4\textsuperscript{th} Jian Pu$^*$}
\IEEEauthorblockA{\textit{Institute of Science and Technology for Brain-Inspired Intelligence} \\
\textit{Fudan University}\\
Shanghai, China \\
jianpu@fudan.edu.cn}

\and

\IEEEauthorblockN{5\textsuperscript{th} Xiangyang Xue}
\IEEEauthorblockA{\textit{School of Computer Science} \\
\textit{Fudan University}\\
Shanghai, China \\
xyxue@fudan.edu.cn}}

\maketitle

\begin{abstract}
Monocular 3D object detection aims to locate  objects in different scenes with just a single image. Due to the absence of depth information, several monocular 3D detection techniques have emerged that rely on auxiliary depth maps from the depth estimation task. There are multiple approaches to understanding the representation of depth maps, including treating them as pseudo-LiDAR point clouds, leveraging implicit end-to-end learning of depth information, or considering them as an image input. However, these methods have certain drawbacks, such as their reliance on the accuracy of estimated depth maps and suboptimal utilization of depth maps due to their image-based nature. While LiDAR-based methods and convolutional neural networks (CNNs) can be utilized for pseudo point clouds and depth maps, respectively, it is always an alternative. In this paper, we propose a framework named the Adaptive Distance Interval Separation Network (ADISN) that adopts a novel perspective on understanding depth maps, as a form that lies between LiDAR and images. We utilize an adaptive separation approach that partitions the depth map into various subgraphs based on distance and treats each of these subgraphs as an individual image for feature extraction. After adaptive separations, each subgraph solely contains pixels within a learned interval range. If there is a truncated object within this range, an evident curved edge will appear, which we can leverage for texture extraction using CNNs to obtain rich depth information in pixels. Meanwhile, to mitigate the inaccuracy of depth estimation, we designed an uncertainty module. To take advantage of both images and depth maps, we use different branches to learn localization detection tasks and appearance tasks separately. Our approach significantly enhances the baseline and outperforms depth-assisted techniques, as shown by our extensive experiments on the KITTI monocular 3D object detection benchmark.
\end{abstract}


\section{Introduction}

The urgent need for inexpensive sensors to perceive the 3D world in real-world applications, such as autonomous driving, virtual reality, and robotics, is the motivation for this study. LiDAR devices have shown promising results \cite{b42,b43}; however, their high cost makes them challenging to deploy and maintain widely. To address this issue, there is a pressing need for cheaper object detectors based on vision. Monocular 3D object detection, which uses only one camera to perceive object localization, physical dimensions, and orientation, has received increasing attention from both industry and academia. Despite the potential benefits, accurate 3D information perception remains a significant challenge due to the lack of depth information available from monocular cameras. Therefore, there is a need to solve several problems in monocular 3D object detection.
\begin{figure}[t]
\centerline{\includegraphics[width=1\linewidth]{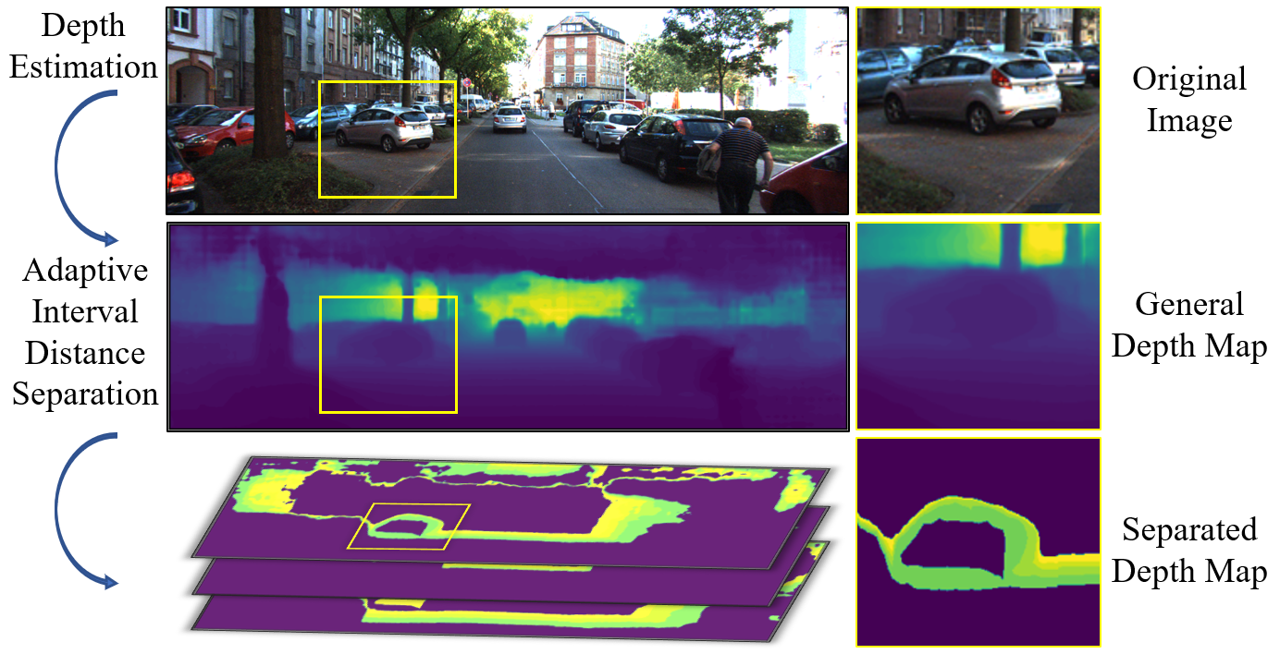}}
\caption{Comparison between general depth maps and our proposed separated depth maps is shown in the figure. The top image shows the original input, the middle image displays the estimated depth map, and the bottom image illustrates our separated depth map. Notably, in the yellow box, a parked car is barely visible in the general depth map but is clearly distinguishable in our separated depth map.}
\label{fig1}
\end{figure}
Recent methods of monocular 3D object detection can be categorized into four main groups based on their ideas for improvement: image-based, imitation-based, pseudo LiDAR-based, and depth-assisted approaches.

Image-based methods rely solely on an image as input. Given the absence of depth information, these methods attempt to establish connections between 2D and 3D object information using geometric methods or feature enhancement techniques. Unlike LiDAR, which can accurately measure distance, monocular methods face challenges in accurately locating objects in space based on the image plane.

Pseudo-LiDAR-based methods aim to bridge the gap between pixels and point clouds. Depth maps from the depth estimation task can be transformed to generate pseudo-LiDAR point clouds. Well-developed LiDAR-based detectors can then be directly applied, while some methods use CNNs to process pseudo-LiDAR. However, the accuracy of depth estimation methods may significantly affect the data transformation of pseudo-LiDAR, making it unsuitable for point cloud detectors trained on accurate point cloud data. Furthermore, overfitting frequently occurs, particularly in the KITTI dataset \cite{b35}.

To achieve excellent performance similar to LiDAR-based or multicamera image-based methods, knowledge distillation is used to improve monocular 3D object detectors. Successful teacher models impressively guide monocular detectors. However, the strict requirements for sensor space-time synchronization and calibration make it challenging to deploy the model on different vehicles. 

Depth-assisted methods complement the rich semantic information of images with the distance information of depth maps. Previous works mainly focus on effectively extracting and fusing depth information with image features, often using CNNs. CNNs are suitable for extracting semantic and texture features, but depth maps lack these features. Therefore, the decision to let depth maps directly pass through CNNs needs reconsideration. Additionally, the accuracy of the depth map estimation significantly affects the model's performance. It is essential to distinguish which parts of the depth map are trustworthy and should be given special consideration. Furthermore, images and depth maps have different areas of expertise in the 3D object detection task. Images are better suited for extracting semantic information, while depth maps are ideal for determining location information. Therefore, they should play to their strengths and complement each other.

In this paper, we summarize three problems and provide corresponding solutions. The problems are as follows: (i) Extracting depth map information effectively. Depth maps have physical meaning, while CNNs are better suited for texture extraction. To extract depth map information more effectively, we need to make a connection between texture and physical distance. One way to do this is by dividing the depth map into sub-depth maps (SDs) based on the specified distance. We can then preserve pixels falling into a specific range as another sub-depth map. By doing this, we can combine the advantages of depth maps and CNNs, better recognize edges, and have an intuitive explanation. (ii) Dealing with inaccurate depth maps. To address the problem of inaccurate depth map estimation, we propose using an uncertainty map aligned with the size of the depth map. The uncertainty map gives confidence for every pixel on the depth map and allows the model to judge whether the location of depth maps is credible. If the depth value tends to be inaccurate, the corresponding uncertainty map suppresses the error caused by the inaccurate region, and the credible values play a greater role. The module of the uncertainty map generally makes depth maps more reliable. (iii) Utilizing the advantages of images and depth maps. Images and depth maps have their respective advantages in appearance and localization features. To make the most of these advantages, we designed a feature decoupling module that separates output heads according to the input modality. By using different branches to learn appearance and localization detection task heads, the model can better recognize categories and understand spatial information.

Our main contributions are as follows:
\begin{itemize}
\item We actively create edges to connect the physical meaning of depth maps with the ability of CNNs. 
\item We propose a simple and effective uncertainty map to reduce the noise caused by inaccurate depth estimation and make the depth map more reliable.
\item We separate output heads according to the input modality, which improves the overall performance of the model without affecting the calculation cost.
\item Extensive experiments show the effectiveness of our proposed approach among monocular 3D object detection methods on the KITTI benchmark.
\end{itemize}

\begin{figure*}[ht]
\centering
\includegraphics[width=1\linewidth]{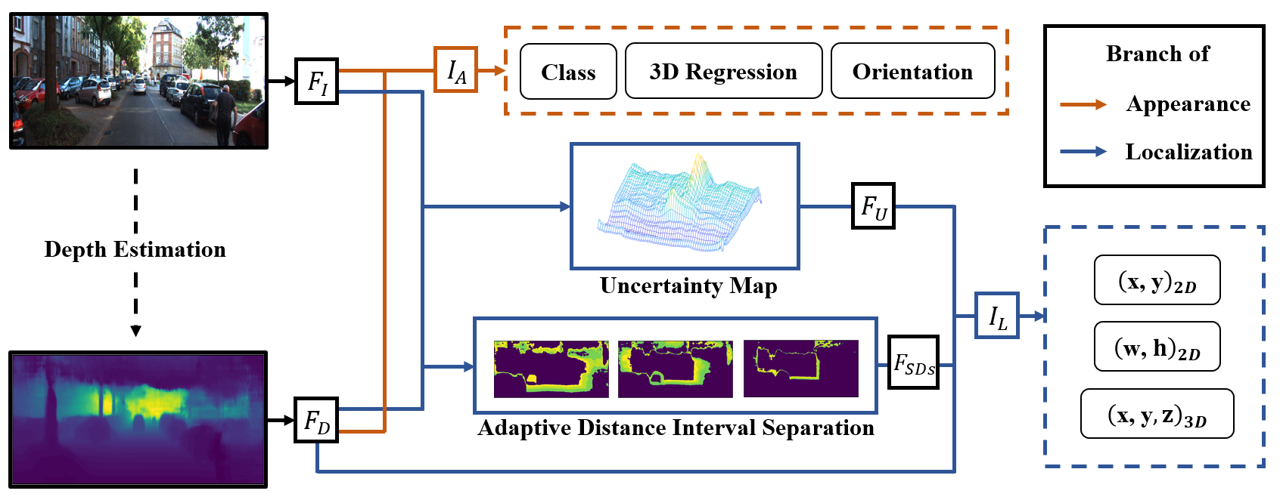}
\caption{
     Visualization of our proposed framework is shown below, where two branches for appearance and localization are denoted in different colors. Dashed boxes represent corresponding output tasks. In the appearance branch, the appearance information $I_A$ is generated from both the image feature $F_I$ and the depth feature $F_D$. The uncertainty map and separated depth maps with adaptive distance intervals are generated by encoding $F_I$ and $F_D$ as $F_U$ and $F_{SDs}$, respectively. These features are then fused into localization information $I_L$ and fed to the corresponding heads.
    }
\label{fig2}
\end{figure*}
\section{Related work}


\subsection{Image-based approaches}

It is challenging to infer 3D spatial information from a single image due to the inherent lack of depth information. Pioneering works \cite{b1,b2,b4,b5} use relatively stable 2D detection results to predict 3D boxes. Some approaches match 2D boxes with predefined 3D shapes for different cars \cite{b3}. Other works focus on disentangling complex interactions between parameters, achieving remarkable progress \cite{b6,b7}. M3D-RPN \cite{b8} proposes a framework that shares 2D-3D features with sets of anchors \cite{b9}, while RTM3D \cite{b10} estimates several projected key points to construct 3D bounding boxes with geometric relationships. Heatmap-based approaches consider objects as points \cite{b11, b12, b13,b14}, significantly reducing inference time. In some cases, auxiliary information support can significantly increase the performance of monocular 3D detectors. Kinematic3d \cite{b15} proposes establishing a kinematic model using continuous frames of images and correcting the object's location in space according to the kinematic model recommendations and regression branches' results. CaDDN \cite{b16} learns the categorical depth distribution from sampled point clouds in the bird's-eye-view (BEV) at the training stage, utilizing point clouds to guide depth features.

\subsection{Pseudo LiDAR-based approaches}
Compared to the success of LiDAR-based methods, monocular 3D object detectors have significant room for improvement. Wang et al. \cite{b17} converted depth map pixels obtained from the depth estimation task into a pseudo-LiDAR point cloud representation, which was then processed using a LiDAR-based detection network \cite{b18}. PatchNet \cite{b19} proposed a novel perspective on the pseudo-LiDAR representation. In addition to using LiDAR-based detectors, simple CNNs can also produce considerable results. Wang et al. \cite{b20} analyzed the impact of 3D object parameters on the results and adopted different strategies based on the degree of influence. NeighborVote \cite{b21} proposes a joint decision-making model based on the pseudo-LiDAR generated from the estimated depth map. The feature points surrounding the proposed object form their own predictions, and these predictions are then combined to form the final output.
\subsection{Imitation-based approaches}
In the task of 3D object detection, LiDAR and stereo modalities have achieved remarkable performance. Some algorithms aim to modify monocular 3D detectors by adopting the paradigms of the aforementioned modalities to enable them to perform well under different modality settings \cite{b22,b23,b24}. DD3D \cite{b25} explores the role of depth maps in the detection task and proposes a transfer model between depth estimation and 3D detection. The well-trained knowledge from depth estimation is then applied to the image, and the results demonstrate that the implicit use of depth information can significantly enhance the 3D detection results. CMKD \cite{b26} transfers knowledge from LiDAR to the image via a knowledge distillation semi-supervised training framework, resulting in significant performance improvement. MonoPS \cite{b27} contends that the gap between stereo and image is much smaller than that between LiDAR and image. It replaces stereo images from the stereo-based detection architecture with defined pseudo-stereo image features, resulting in promising results. To imitate the representations of 3D point clouds, Feng et al. \cite{b28} align 2D proposals with 3D point clouds.
\subsection{Depth-assisted approaches}
D4LCN \cite{b29} uses depth maps as guidance to distinguish foreground and background in scenes, while DDMP-3D \cite{b30} utilizes graphic message passing mechanisms and convolution kernels on depth maps to dynamically extract features. Moreover, instance-level center-aware depth guidance is proposed to reduce depth estimation errors. DRF-Net \cite{b31} discusses the appearance and location information of objects and uses a feature reflecting module to make the most of different features. Cai et al. \cite{b32} treat monocular 3D object detection as several sub-tasks, roughly estimating object depth with an auxiliary prior 2D structured polygon and finally refining it with the depth map.

\section{Methodology}
In this section, we present the framework of our proposed ADISN. Subsequently, we provide a detailed description of three major structures proposed in this study, which include adaptive distance interval separation, uncertainty map, and feature decoupling module.

\subsection{Pipeline overview}\label{AA}


We employed M3D-RPN \cite{b8} as our baseline detector due to its simplicity and influence on several depth-assisted methods. We followed the baseline's anchor definition and learning parameters. Our network accepts both image and depth map inputs. We used DORN \cite{b33} as a monocular depth estimator to generate depth maps. As illustrated in Fig \ref{fig2}, we encoded the images and depth maps as $F_I$ and $F_D$, respectively. We fused $F_I$ and $F_D$ to create an uncertainty map $U$ and adaptive distance lower and upper bounds. We extracted feature $F_{SDs}$ from Sub-Depth maps (SDs) and combined it with $F_D$ and $F_U$ from the uncertainty map to obtain the overall localization information feature $I_L$. Simultaneously, we fused $F_I$ and $F_D$ to produce the overall appearance information $I_A$. Finally, $I_L$ and $I_A$ were fed into output task heads with feature decoupling. We elaborate on all the modules mentioned above in the following subsections.
\begin{figure}[t]
\centerline{\includegraphics[width=1\linewidth]{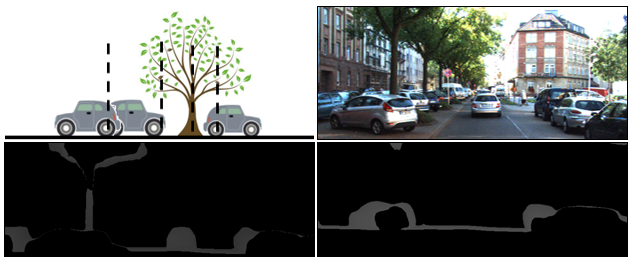}}
\caption{Example of separated sub-depth maps: The top-left panel shows a sketch map of the side view with objects truncated using our adaptive intervals. The top-right panel displays the original input image, while the bottom panels show the separated sub-depth maps.}
\label{fig3}
\end{figure}

\subsection{Adaptive distance interval separation}
In this section, we propose an adaptive distance interval separation module (ADIS) to handle different distance intervals in the depth map. Since the intensity of the pixel value of the depth map represents the physical meaning of distance, we divide the depth map into several intervals based on distance. For each sub-depth (SD), only pixels within the selected distance interval are retained while the remaining positions are assigned to zero. This generates a sharp edge near the reserved area on the subgraph, which is favorable for the strength of CNNs. If the selected distance interval truncates the object, there will be significant changes in the subgraph, making it conducive to positioning the object in space.

We need to generate upper and lower bounds adaptively from each unique input feature to obtain an adaptive distance interval. We define the interval sequence as $\mathbf{d_i}=(d_{i-1},d_i), i=1,2,...,n_d$, where $n_d$ denotes the number of distance intervals we want, and $d_i$ is the distance learned from the network. Every $\mathbf{d_i}$ has its upper and lower bounds, which describe a range of distances. There is no intersection between intervals ($\mathbf{d_i} \cap \mathbf{d_j} = \phi , i \neq j$).

Given feature $F_{ID} \in \mathbb{R}^{C \times H \times W}$, which denotes the fusion of $F_I$ and $F_D$, we use a $1 \times 1$ convolution to reduce the channel and flatten the feature map to a vector of size $1 \times (H \times W)$. After fully connected layers, we have a distance feature of size $1 \times n_d$, and the $\mathbf{d}=(d_0, d_1, ... d_{n_d}), d_0=0$ is generated using the softmax operation. Therefore, we can divide the input depth map $Dep$ into its subgraphs $SD_i, i=1,2,...,n_d$ with $d_i$ using Equation \ref{SDs}.

\begin{equation}
\begin{footnotesize} 
SD_i = \left\{\begin{matrix}
SD^{(w,h)} = Dep^{(w,h)} , & \sum_{j=0}^{i-1}d_j\leq SD^{(w,h)}\leq \sum_{j=0}^{i}d_j  \\ 
SD^{(w,h)} = 0 , & Otherwise.
\end{matrix}
\right.
\label{SDs}
\end{footnotesize} 
\end{equation}

\begin{figure}[t]
\centerline{\includegraphics[width=1\linewidth]{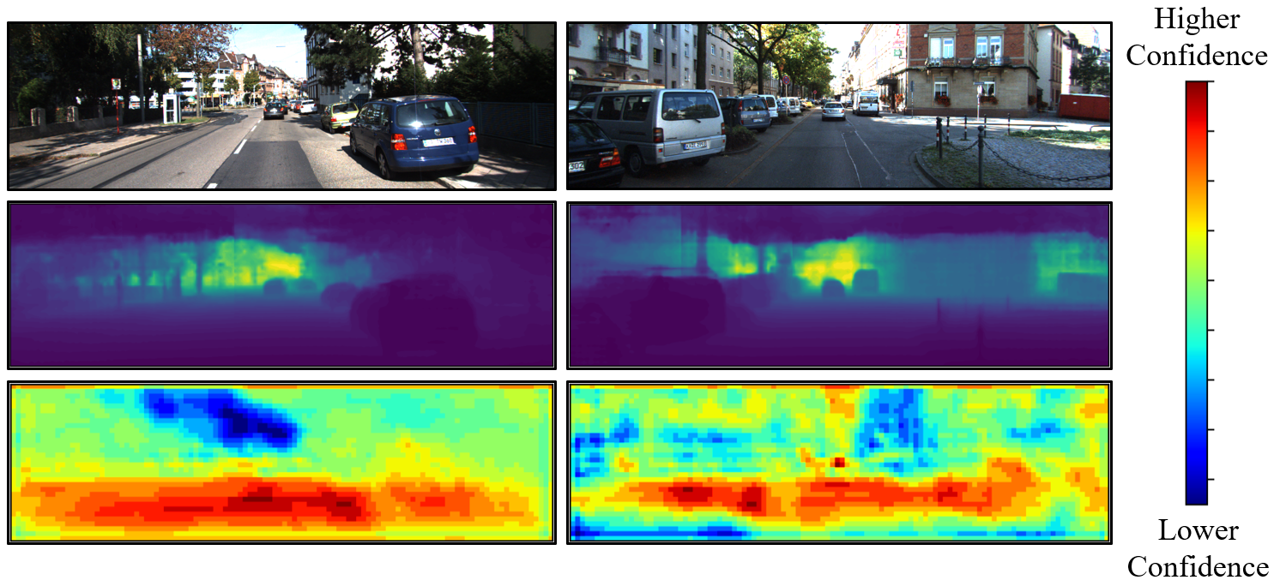}}
\caption{Visualization of uncertainty maps. The top shows input images, the middle displays estimated depth maps, and the bottom exhibits our proposed uncertainty maps. As demonstrated by the confidence levels, nearby areas of the depth maps are more reliable, while the inaccuracy of the distant areas and the sky is reduced.}
\label{fig4}
\end{figure}

Here, $(w,h)$ represents the pixel coordinates in the image of size $H \times W$. The estimated depth value of the pixels is preserved in the SD, and the pixel coordinates correspond to the image, so it benefits CNNs to extract depth information. The ADIS module is more effective than directly applying CNNs to the depth map since it generates SDs with obvious curved edges precisely caused by the truncation of objects. When $n_d$ is large enough, in fact, the SDs we divide would be the dense LiDAR point cloud. Fig \ref{fig3} shows the generated SDs, these SDs  contain obvious curved edges, which are precisely caused by the truncation of objects.


\subsection{Uncertainty map}

Our proposed model heavily relies on the accuracy of the depth map during the feature extraction stage. However, the depth map we utilize is obtained from a pre-existing depth estimation method, which implies that it may not be entirely precise. The regions that are not trained thoroughly by the LiDAR in the depth estimation task, such as faraway objects or the sky, tend to perform poorly, and this inaccuracy remains on the depth map. This can be disadvantageous when estimating the depth of a dense image, as it can negatively impact the model's performance. To address this, we draw inspiration from methods based on uncertainty that utilize uncertainty measurements to mitigate errors when there is noise in the initial condition. Thus, we incorporate an uncertainty map into our network to guide the model during the process of extracting depth information. As we aim to mitigate the inaccuracy of the depth map, our depth extraction module learns the uncertainty point by point, suppressing inaccurate pixels in the estimated depth map and reducing the negative impact on the model. To enable the uncertainty map to locate the regions to be suppressed in the image and depth map, we jointly generate the uncertainty feature $U$ by combining the image and depth map features. This uncertainty feature is also used when extracting $\mathbf{d}$, ensuring that the two modules work together to further improve the accuracy of the depth map. The details of our approach are provided below:
\begin{equation}
U^{H \times W} = \mathbf{1}^{H \times W}-Sigmoid(Upsampling(F_{ID}))^{H \times W},
\label{Un}
\end{equation}
shows the process of generating the uncertainty map, where $\mathbf{1}^{H \times W}$ represents the $H \times W$ matrix with elements of 1, and $F_{ID}$ is the fusion of $F_I$ and $F_D$. The sigmoid function used ensures that each pixel in the corresponding depth map is assigned an uncertainty value between 0 and 1. To avoid gradient vanishing in the early stage of training, we obtain $1-U$ instead of directly predicting $U$. The trained uncertainty maps are presented in \ref{fig4}. The final output feature is the result of combining the SDs, which is obtained through the following equation:
\begin{equation}
F_{SD} = Conv(\mathbf{SD} \times \mathbf{U}),
\end{equation}
where $\mathbf{SD}$ and $\mathbf{U}$ denote the stacked $SD_i$ and $U$, respectively. Both are duplicated to the same channel according to the $n_d$.

\begin{table*}[t]
\renewcommand{\arraystretch}{1.2}
\caption{
Comparison of Different Methods for the ``Car" Category on the KITTI Test Set with AP 3D and AP BEV. We Highlight the Best Results in \textbf{bold} and the Second Place in \underline{underlined}}
\label{table1}
\begin{center}
\resizebox{.8\textwidth}{!}{
\begin{tabular}{|c|c|ccc|ccc|}
\hline
\multirow{2}{*}{Experiments} &
  \multirow{2}{*}{Implementation mode} &
  \multicolumn{3}{c|}{AP 3D} &
  \multicolumn{3}{c|}{AP BEV} \\ \cline{3-8} 
 &
   &
  \multicolumn{1}{c|}{Easy} &
  \multicolumn{1}{c|}{Mod.} &
  Hard &
  \multicolumn{1}{c|}{Easy} &
  \multicolumn{1}{c|}{Mod.} &
  Hard \\ \hline
M3D-RPN(base line) \cite{b8} &
  image-based &
  \multicolumn{1}{c|}{14.76} &
  \multicolumn{1}{c|}{9.71} &
  7.42 &
  \multicolumn{1}{c|}{21.02} &
  \multicolumn{1}{c|}{13.67} &
  10.23 \\ \hline
MonoCInIS \cite{b36} &
  image-based &
  \multicolumn{1}{c|}{15.82} &
  \multicolumn{1}{c|}{7.94} &
  6.68 &
  \multicolumn{1}{c|}{22.28} &
  \multicolumn{1}{c|}{11.64} &
  9.95 \\ \hline
ImVoxelNet\cite{b37} &
  image-based &
  \multicolumn{1}{c|}{17.15} &
  \multicolumn{1}{c|}{10.97} &
  9.15 &
  \multicolumn{1}{c|}{25.19} &
  \multicolumn{1}{c|}{16.37} &
  13.58 \\ \hline
M3DSSD \cite{b38}&
  image-based &
  \multicolumn{1}{c|}{\underline{17.51}} &
  \multicolumn{1}{c|}{11.46} &
  8.98 &
  \multicolumn{1}{c|}{24.15} &
  \multicolumn{1}{c|}{15.93} &
  12.11 \\ \hline
AM3D \cite{b39}&
  pseudo LiDAR-based &
  \multicolumn{1}{c|}{16.50} &
  \multicolumn{1}{c|}{10.74} &
  9.52 &
  \multicolumn{1}{c|}{25.03} &
  \multicolumn{1}{c|}{17.32} &
  14.91 \\ \hline
PatchNet\cite{b19} &
  pseudo LiDAR-based &
  \multicolumn{1}{c|}{15.68} &
  \multicolumn{1}{c|}{11.12} &
  \underline{10.17} &
  \multicolumn{1}{c|}{22.97} &
  \multicolumn{1}{c|}{16.86} &
  14.97 \\ \hline
Zhao et al.\cite{b40} &
  pseudo LiDAR-based &
  \multicolumn{1}{c|}{15.86} &
  \multicolumn{1}{c|}{11.41} &
  10.08 &
  \multicolumn{1}{c|}{\textbf{28.79}} &
  \multicolumn{1}{c|}{\textbf{18.61}} &
  \underline{15.27} \\ \hline
Cai et al.\cite{b32} &
  depth-assisted &
  \multicolumn{1}{c|}{11.68} &
  \multicolumn{1}{c|}{7.28} &
  5.69 &
  \multicolumn{1}{c|}{24.62} &
  \multicolumn{1}{c|}{14.66} &
  11.46 \\ \hline
Ji et al.\cite{b41} &
  depth-assisted &
  \multicolumn{1}{c|}{12.46} &
  \multicolumn{1}{c|}{7.86} &
  6.30 &
  \multicolumn{1}{c|}{19.85} &
  \multicolumn{1}{c|}{13.07} &
  10.29 \\ \hline
D4LCN \cite{b29} &
  depth-assisted &
  \multicolumn{1}{c|}{16.65} &
  \multicolumn{1}{c|}{\underline{11.72}} &
  9.51 &
  \multicolumn{1}{c|}{22.51} &
  \multicolumn{1}{c|}{16.02} &
  12.55 \\ \hline
ADISN(Ours) &
  depth-assisted &
  \multicolumn{1}{c|}{\textbf{17.87}} &
  \multicolumn{1}{c|}{\textbf{12.14}} &
  \textbf{10.42} &
  \multicolumn{1}{c|}{\underline{27.18}} &
  \multicolumn{1}{c|}{\underline{17.91}} &
  \textbf{15.33} \\ \hline
\end{tabular}}
\end{center}
\end{table*}

\subsection{Feature decoupling}
Inspired by DRF-Net \cite{b31}, DDMP-3D \cite{b30}, and other monocular 3D object detection methods, it has been noted that the position of the object in space, i.e., $(x, y, z)$, is the most critical factor in the performance of monocular 3D detection. However, images suffer from the weakness of understanding depth, and the depth estimation task exists to address this issue. Experiments have shown that even if only a single path of depth maps is used as input, the performance of 3D detection is better than when only RGB images are used. Nevertheless, relying solely on depth maps makes it difficult to understand the appearance and category of the object. Moreover, over-reliance on depth maps is affected by inaccurate depth estimation, especially at long distances. Therefore, RGB images are still necessary. They perform better in semantics than depth maps and can complement each other, which is also the advantage of depth-assisted methods. From this perspective, appearance and location features should be different and related. Compared with DRF-Net \cite{b31} from the perspective of general features, we consider the modality of input and use ADIS for further learning of depth maps.

Specifically, we divide the detection head into two streams: the appearance head and the localization head. The feature before the final output $I_A$ represents the appearance information, while $I_L$ represents the location information, each generated in the previous stage, $I_A = F_I + F_D$, and $I_L = F_D+F_{SD}+F_U$. In the stream of appearance, we still maintain the depth features. By robustly combining the extracted RGB and depth features, our algorithm does not rely too much on the depth map that may be biased, thus improving the overall performance of our model.

\section{Experiments}

\subsection{Dataset}

The KITTI dataset is a widely recognized benchmark for evaluating 3D object detection algorithms, comprising 7481 training samples and 7518 testing samples, with both 3D point cloud and color images provided. In our approach, we rely solely on color images. The evaluation is conducted on three categories of objects, namely Car, Pedestrian, and Cyclist, using the average precision (AP) metric, which is a commonly used evaluation metric for 3D and Bird's-Eye-View(BEV) object detection, calculated using 40 recall positions. As per the official KITTI evaluation protocol, objects are classified into three difficulty levels: easy, moderate, and hard based on the object size, occlusion state, and truncation level. The algorithms' final performance is ranked based on the moderately difficult results. In this paper, we use the Average Precision (AP) as the evaluation metric, with an Intersection over Union (IoU) threshold of 0.7. Under this standard, the definition of AP is as follows: for each class, we calculate a set of confidence scores and corresponding ground-truth labels, and sort them in descending order of confidence scores. Then, for each confidence threshold, we compute the precision and recall, and the corresponding AP value. Finally, we take the average AP over all classes to obtain the final AP value. This metric can be used to evaluate the accuracy and robustness of object detection models.

\subsection{Implementation details}
We implemented our proposed method using PyTorch 1.2.0. Ablation study experiments were conducted on a GTX1080Ti GPU with 11 GB memory, with a batch size of 2 and a learning rate of 0.004 for 80000 iterations. For the test set, we trained with a batch size of 4, a learning rate of 0.008 for 120000 iterations, and a Tesla V100 with 32 GB. We used DenseNet-121 as the backbone for extracting both the image and depth map. Since the backbone is trained on ImageNet, we normalized the input using means of (0.485, 0.456, 0.406) and stds of (0.229, 0.224, 0.225), and the optimizer was SGD. We set the number of intervals $n_d$ to 8. Since $n_d$ is generated using full connection layers, the size of the feature map must be strictly controlled, hence we padded every input image to size (1760, 512) at the beginning of both the training and testing steps. 

\begin{table}[ht]
\renewcommand{\arraystretch}{1.2}
\caption{
Comparison of the Use of Different Depth Maps for Category ``Car" on the KITTI Validation Set with AP 3D}
\label{table3}
\begin{center}
\resizebox{1\linewidth}{!}{
\begin{tabular}{|c|c|ccc|}
\hline
\multirow{2}{*}{Experiments} & \multirow{2}{*}{Depth map} & \multicolumn{3}{c|}{AP 3D}                                       \\ \cline{3-5} 
                             &                            & \multicolumn{1}{c|}{Easy}   & \multicolumn{1}{c|}{Mod.}  & Hard  \\ \hline
M3D-RPN(base line) \cite{b8}           & -                          & \multicolumn{1}{c|}{14.53}  & \multicolumn{1}{c|}{11.07} & 8.65  \\ \hline
D4LCN \cite{b29}                        & DORN\cite{b33}                       & \multicolumn{1}{c|}{22.32}  & \multicolumn{1}{c|}{16.20} & 12.30 \\ \hline
ADISN(Ours) & DORN\cite{b33}   & \multicolumn{1}{c|}{\textbf{26.86}} & \multicolumn{1}{c|}{\textbf{17.28}} & \textbf{13.15} \\
Improvement over \cite{b29}          & -                          & \multicolumn{1}{c|}{+4.54}  & \multicolumn{1}{c|}{+1.08} & +0.85 \\ \hline
D4LCN \cite{b29}                        & PSMNet\cite{b34}                     & \multicolumn{1}{c|}{25.24}  & \multicolumn{1}{c|}{19.80} & 16.45 \\ \hline
ADISN(Ours) & PSMNet\cite{b34} & \multicolumn{1}{c|}{\textbf{36.02}} & \multicolumn{1}{c|}{\textbf{23.25}} & \textbf{18.19} \\
Improvement over \cite{b29}          & -                          & \multicolumn{1}{c|}{+10.78} & \multicolumn{1}{c|}{+3.45} & +1.74 \\ \hline
\end{tabular}}
\end{center}
\end{table}

\subsection{Comparison}
Table \ref{table1} displays the performance comparison of the KITTI test set in the category of cars based on AP 3D and AP BEV, compared under the IoU=0.7 standard. Our proposed method achieves superior performance among monocular 3D object detection methods, ranking 1$^{st}$ in the Easy, Moderate, and Hard categories with AP 3D. We observe significant improvement in the 3D detection task compared to other advanced methods, with an increase of 3.11 on Easy, 2.43 on Moderate, and 3.00 on Hard when compared to the baseline. Our method is based on M3D-RPN \cite{b8}, and despite the improved M3D-RPN \cite{b8} named ``3D-Net" being used as the baseline by D4LCN \cite{b29}, our method retains its advantages. This indicates that the depth map has a significant improvement in the monocular 3D object detection task, and that older methods such as the baseline are still competitive in recent methods. Under the AP BEV index, we rank second in Easy and Moderate and first in Hard. Recently, monocular methods no longer compare results on the validation set, as performance often drops significantly on the test set. Therefore, comparing results on the test set is the most objective method. Overall, our comprehensive experimental results confirm that our method significantly improves the monocular 3D object detector.

\begin{table}[t]
\renewcommand{\arraystretch}{1.2}
\caption{Comparison of Using Different $n_d$ for Category ``Car" on the KITTI Validation Set}
\label{table4}
\begin{center}
\resizebox{.6\linewidth}{!}{
\begin{tabular}{|c|ccc|}
\hline
\multirow{2}{*}{Settings of $n_d$} & \multicolumn{3}{c|}{AP 3D}                                                           \\ \cline{2-4} 
                                   & \multicolumn{1}{c|}{Easy}  & \multicolumn{1}{c|}{Mod.}  & Hard                       \\ \hline
0                                  & \multicolumn{1}{l|}{22.93} & \multicolumn{1}{l|}{14.81} & \multicolumn{1}{l|}{11.58} \\ \hline
4                                  & \multicolumn{1}{c|}{23.64} & \multicolumn{1}{c|}{16.21} & 11.47                      \\ \hline
8                                  & \multicolumn{1}{c|}{26.86} & \multicolumn{1}{c|}{\textbf{17.28}} & \textbf{13.15}                      \\ \hline
16                                 & \multicolumn{1}{c|}{\textbf{27.92}} & \multicolumn{1}{c|}{17.14} & 13.15                      \\ \hline
32                                 & \multicolumn{1}{c|}{26.65} & \multicolumn{1}{c|}{16.68} & 12.64                      \\ \hline
\end{tabular}}
\end{center}
\end{table}

\begin{table*}[ht]
\renewcommand{\arraystretch}{1.2}
\caption{
Comparison of Various Modules in the Proposed Method for the Category ``Car" on the KITTI Validation Set}
\label{table2}
\begin{center}
\resizebox{.9\linewidth}{!}{
\begin{tabular}{|c|cccc|ccc|}
\hline
\multirow{2}{*}{Experiments} &
  \multicolumn{4}{c|}{Our proposed operations} &
  \multicolumn{3}{c|}{AP BEV} \\ \cline{2-8} 
 &
  \multicolumn{1}{c|}{Separated head} &
  \multicolumn{1}{c|}{ADIS(8)} &
  \multicolumn{1}{c|}{Un} &
  Feature decoupling &
  \multicolumn{1}{c|}{Easy} &
  \multicolumn{1}{c|}{Mod.} &
  Hard \\ \hline
M3D-RPN(image only)\cite{b8} &
  \multicolumn{1}{c|}{} &
  \multicolumn{1}{c|}{} &
  \multicolumn{1}{c|}{} &
   &
  \multicolumn{1}{c|}{14.43} &
  \multicolumn{1}{c|}{10.94} &
  8.57 \\ \hline
M3D-RPN(depth only)\cite{b8} &
  \multicolumn{1}{c|}{} &
  \multicolumn{1}{c|}{} &
  \multicolumn{1}{c|}{} &
   &
  \multicolumn{1}{c|}{19.75} &
  \multicolumn{1}{c|}{13.08} &
  9.46 \\ \hline
M3D-RPN(image+depth)\cite{b8} &
  \multicolumn{1}{c|}{} &
  \multicolumn{1}{c|}{} &
  \multicolumn{1}{c|}{} &
   &
  \multicolumn{1}{c|}{20.90} &
  \multicolumn{1}{c|}{14.33} &
  9.93 \\ \hline
\multirow{5}{*}{ours} &
  \multicolumn{1}{c|}{\ding{52}} &
  \multicolumn{1}{c|}{} &
  \multicolumn{1}{c|}{} &
   &
  \multicolumn{1}{c|}{22.93} &
  \multicolumn{1}{c|}{14.81} &
  11.58 \\ \cline{2-8} 
 &
  \multicolumn{1}{c|}{\ding{52}} &
  \multicolumn{1}{c|}{\ding{52}} &
  \multicolumn{1}{c|}{} &
   &
  \multicolumn{1}{c|}{23.09} &
  \multicolumn{1}{c|}{15.77} &
  12.15 \\ \cline{2-8} 
 &
  \multicolumn{1}{c|}{\ding{52}} &
  \multicolumn{1}{c|}{\ding{52}} &
  \multicolumn{1}{c|}{\ding{52}} &
   &
  \multicolumn{1}{c|}{23.84} &
  \multicolumn{1}{c|}{16.12} &
  12.64 \\ \cline{2-8} 
 &
  \multicolumn{1}{c|}{\ding{52}} &
  \multicolumn{1}{c|}{\ding{52}} &
  \multicolumn{1}{c|}{} &
  \ding{52} &
  \multicolumn{1}{c|}{25.03} &
  \multicolumn{1}{c|}{16.63} &
  12.43 \\ \cline{2-8} 
 &
  \multicolumn{1}{c|}{\ding{52}} &
  \multicolumn{1}{c|}{\ding{52}} &
  \multicolumn{1}{c|}{\ding{52}} &
  \ding{52} &
  \multicolumn{1}{c|}{\textbf{26.86}} &
  \multicolumn{1}{c|}{\textbf{17.28}} &
  \textbf{13.15} \\ \hline
\end{tabular}}
\end{center}
\end{table*}

\subsection{Ablation Study}

We compared our proposed method with M3D-RPN \cite{b8} and D4LCN \cite{b29} on the KITTI validation set using different depth estimation methods (see Table \ref{table3}). Our method showed advantages with different depth maps, with greater advantages when the depth map accuracy was higher. This suggests that our method's approach to understanding depth maps is more effective, as a more accurate depth estimation is closer to point clouds and more intuitive.

We conducted a simple experiment to determine the number of intervals ($n_d$) according to the final model, as shown in Table \ref{table4}. Increasing $n_d$ improves the effect, but also incurs greater computational costs. When $n_d$ was set to 16, performance on Hard no longer increased, as hard objects are difficult to estimate accurately in the depth estimation task. Setting $n_d$ to greater than 32 may lead to overfitting and degrade performance. After comprehensive consideration, we set $n_d$ to 8.

Table \ref{table2} shows the ablation experiment of our proposed method. We disassembled each module to verify their effectiveness. The performance of M3D-RPN \cite{b8} as a baseline was lower than the original paper's report due to device issues. Notably, even using a depth map directly as the input of M3D-RPN \cite{b8} yielded better performance than the original RGB image, indicating the crucial role of depth maps in extracting position information for monocular 3D object detection. A simple fusion of image and depth map features yielded performance improvements, likely due to the complementarity of depth maps and images. We separately set the detector heads of different tasks according to feature characteristics to improve the results. The ADIS module significantly improved performance by further extracting depth features. Adding an uncertainty map also improved overall performance. Furthermore, our feature decoupling module further improved performance.

\begin{figure*}[ht]
\centering
\includegraphics[width=0.82\linewidth]{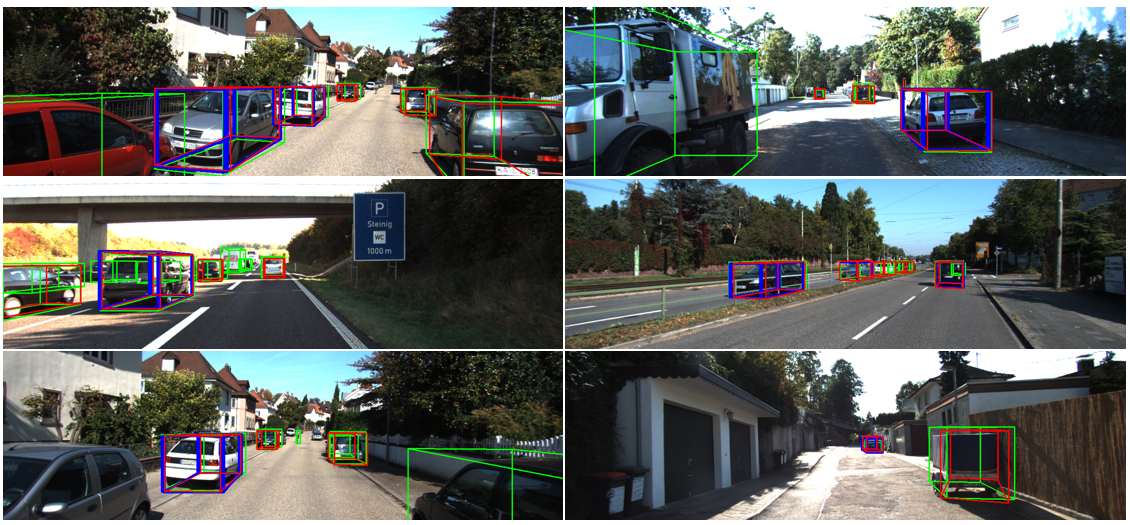}
\caption{The qualitative comparison of the ground truth (green), the baseline (blue), and our proposed method (red) on the KITTI validation set. Our proposed method predicts accurate 3D bounding boxes of occluded and distant objects.}
\label{fig5}
\end{figure*}

Fig \ref{fig5} shows visualizations of the effectiveness of our proposed method. Our method outperformed M3D-RPN \cite{b8} for distant or occluded objects and improved many missed detection situations.

\newpage

\section{Conclusion}

In this paper, we present a novel approach to understanding depth map representation by proposing ADISN. Our sub-depth map establishes a relationship between 3D and 2D expressions, providing insights into the physical meaning of depth maps and highlighting the advantages of CNNs over converting depth maps to 3D or 2D representations. To address the issue of inaccurate estimated depth maps, we introduce the use of an uncertainty map, which improves the reliability of learning in unclear areas by relying more on clear images. Additionally, we propose a method where appearance and localization heads learn from different branches, allowing images and depth maps to leverage their respective strengths. Extensive experimentation shows that our proposed method outperforms the highly competitive KITTI monocular 3D object detection task.

\section{Future work}

Monocular 3D object detection is always a hot topic in the field of autonomous driving before reliable and cost-effective LiDAR sensors become available. Recent research has mostly focused on improving the performance of monocular 3D object detectors using other modalities, partly because it is relatively easy to adapt existing high-performance algorithms by pruning them into monocular detectors, and partly because it may be inappropriate to rely solely on a single image to reconstruct a 3D scene. Previous work has pre-trained neural networks on depth estimation tasks and fine-tuned them on object detection tasks with color images, implicitly learning depth features, and achieving impressive performance. Modalities or paradigms that are independent of RGB inputs may possess more knowledge suitable for inferring 3D space, and extracting and refining such knowledge into single images may be a direction for future work.




\vspace{12pt}
\color{red}

\end{document}